\documentclass[letterpaper]{article} 
\usepackage{aaai2027}
\copyrighttext{Preprint}
\usepackage[hyphens]{url}  
\usepackage{graphicx} 
\def\UrlFont{\rm}  
\usepackage{natbib}  
\usepackage{caption} 
\usepackage{algorithm}
\usepackage{algorithmic}

\usepackage{newfloat}
\usepackage{listings}
\DeclareCaptionStyle{ruled}{labelfont=normalfont,labelsep=colon,strut=off} 
\floatstyle{ruled}
\newfloat{listing}{tb}{lst}{}
\floatname{listing}{Listing}

\usepackage{booktabs}

\title{Logic-Guided Graph Denoising Diffusion Models}
\author{
    Samuel Cognolato\textsuperscript{\rm 1,\rm 2},
    Luciano Serafini\textsuperscript{\rm 2},
    Alessandro Sperduti\textsuperscript{\rm 1,\rm 2,\rm 3},
    Andrea Cavallaro\textsuperscript{\rm 4}
}
\affiliations{
    \textsuperscript{\rm 1}Department of Mathematics ``Tullio Levi-Civita'', University of Padova, Padova, Italy\\
    \textsuperscript{\rm 2}Fondazione Bruno Kessler (FBK), Trento, Italy\\
    \textsuperscript{\rm 3}Human Inspired Technology Research Centre (HIT), University of Padova, Padova, Italy\\
    \textsuperscript{\rm 4}\'Ecole polytechnique f\'ed\'erale de Lausanne (EPFL), Lausanne, Switzerland
}

\usepackage{amsmath}
\usepackage{amsfonts}
\usepackage{booktabs}
\usepackage{multirow}
\usepackage{makecell}
\usepackage{pifont}
\usepackage{subcaption}

\usepackage{bm}

\newcommand{\cmark}{\ding{51}}
\newcommand{\xmark}{\ding{55}}
\newcommand{\meanstd}[2]{#1\,$\pm$\,#2}

\definecolor{white}{rgb}{1,1,1}

\DeclareMathAlphabet{\mathsfit}{\encodingdefault}{\sfdefault}{m}{sl}
\SetMathAlphabet{\mathsfit}{bold}{\encodingdefault}{\sfdefault}{bx}{n}

\def\verts{{\mathcal{V}}}
\def\edges{{\mathcal{E}}}

\def\labels{{\mathcal{L}}}
\def\functions{{\mathcal{F}}}

\def\interp{{\mathcal{I}}}
\def\triples{{\mathcal{T}}}
\def\constants{{\mathcal{C}}}
\def\predicates{{\mathcal{P}}}
\def\domain{{\mathcal{D}}}

\def\relations{{\mathcal{R}}}
\def\semantics{{\mathcal{S}}}

\def\adj{{\mathbf{A}}}  
\def\fX{{\bm{X}}}       
\def\fE{{\bm{E}}}     

\def\fG{{\bm{G}}}       

\def\R{{\mathbb{R}}}    

\def\m1{{\bm{1}}}   

\usepackage{comment}

\newif\ifshowcomments
\showcommentsfalse   

\ifshowcomments
  \newenvironment{mycomment}
    {\color{red}}
    {}
\else
  \excludecomment{mycomment}
\fi

\begin{document}

\maketitle

\begin{abstract}
Graph diffusion models are competitive approaches for generating complex graph-structured data. However, when such distributions have strict structural constraints, diffusion models may fail to capture them exactly, resulting in the generation of invalid graphs. In this paper, we propose a method for grounding graph diffusion models in domain-specific knowledge via First Order Logic (FOL), leveraging its expressive and interpretable formalism to impose structural constraints on the generative process. We incorporate domain knowledge using classifier guidance, a technique that conditions the diffusion model's distribution on external variables to exert control over the sampled data. During the generative process, classifier guidance steers the denoising trajectory to optimize an objective function, which we define as a fuzzy and differentiable relaxation of the logical rules. Our methodology requires only the specification of constraints in FOL, whose satisfaction is encouraged at inference time to guide a pre-trained generative model. We evaluate our approach by comparing the unconditional generation of knowledge graphs with our logic-guided framework, which incorporates domain ontologies as structural constraints during inference. This approach significantly increases the proportion of generated graphs that satisfy the ontological constraints.
\end{abstract}

\begin{links}
\end{links}

\section{Introduction}

Graph generation involves learning an underlying data distribution to facilitate the synthesis of novel, realistic graph structures. Graph diffusion models~\cite{gg_jo2022gdss,gg_vignac2023digress,gg_huang2023cdgs, cognolato2026d4,cognolato2026flagg} and flow-matching models~\cite{gg_eijkelboom2024variational,gg_qin2024defog} can model distributions over graphs with discrete labels and continuous attributes. In some applications, such as biology, graphs need to satisfy domain-specific rules~\cite{rigoni2020conditional} that define whether a graph is well-formed. However, neural generative models lack a mechanism to explicitly enforce structural constraints, which requires specialized techniques to tackle them. Thus, the problem of constrained graph generation, i.e., sampling graphs from a subspace of elements satisfying a set of constraints, has gained traction~\cite{liu2018constrained,DBLP:journals/ml/RigoniNS25,cg_madeira2024generative}.

First Order Logic (FOL)~\cite{barwise1977fol} may serve as a unified approach to encode and combine heterogeneous constraints.
FOL can be used to formalize properties in graphs, where node and edge labels are represented as unary and binary predicates, respectively~\cite{cai1992optimal}. The rules may act on single nodes, their neighbors, and the type or direction of the relations, e.g., bipartiteness and transitivity. FOL rules may also act considering the whole graph structure, e.g., requiring the presence of a specific node type. The ability to operate both locally and globally provides a tool for controlling the structure and content of graphs. This setup has enabled the definition of ontologies for knowledge graphs, providing a schema for otherwise unstructured data~\cite{GRUBER1993199}.

In this paper, we propose LTN-guidance, a methodology for injecting domain knowledge as FOL constraints into the generative process of graph diffusion models~\cite{gg_vignac2023digress}, without any additional training. Our aim is to increase the probability of sampling valid graphs. First, we propose a graph signature, i.e., the collection of symbols defining the vocabulary of the first order language, enabling the definition of logical formulas over labeled graphs. During generation, we use classifier guidance~\cite{dhariwal2021diffusion} to steer a pre-trained diffusion model toward optimizing rule satisfaction (the external objective function). Our method provides the target FOL constraints as Logic Tensor Network (LTN)~\cite{badreddine2022logic} differentiable functions. A mechanism based on Focal Loss~\cite{lin2017focal} allows adaptive reweighting of constraints, increasing the contribution of less satisfied sentences in the generative process. Finally, we propose cheaper implementations of LTN functions, e.g., for implementing cardinality constraints over edges.

Our main contributions are:
\begin{enumerate}
    \item A graph signature that allows the definition of logical rules over labeled graphs. We integrate the signature with axioms to ensure label exclusivity;
    \item A plug-and-play method for steering graph diffusion models with knowledge in FOL. During inference, FOL constraints are translated to differentiable functions through Logic Tensor Network, and used to steer a pretrained unconditional model with classifier guidance.
    \item We evaluated a popular graph diffusion model, DiGress~\cite{gg_vignac2023digress}, with and without our technique, for the generation of small knowledge graphs from the IntelliGraphs benchmark~\cite{thanapalasingam2023intelligraphs} against knowledge graph embedding and LLM-based approaches. We observe an improvement in validity, achieving $>95\%$ in most datasets.
    \item We performed generation with new rules, compatible and conflicting with the data, to evaluate the conditional in-distribution and out-of-distribution validity, respectively. LTN-guidance consistently shows responsiveness to input formulas by producing a high fraction of valid graphs.
\end{enumerate}
To the best of our knowledge, this is the first inference-time FOL constraints guidance framework for graph diffusion.
\section{Related Work}\label{sec:related}

\begin{mycomment}
\begin{table*}[t]
    \fontsize{7}{9}\selectfont
    \setlength{\tabcolsep}{3pt} 
    \centering
    \caption{Methods for constrained graph generation. We report the type of contraints, defining the domain in which the method is applicable, the generative model in use, whether it is applicable to First Order Logic, and if it is, the theoretical maximum number of variables for which it can be used.}
    \begin{tabular}{@{}cc*{4}{c}@{}}
        \toprule
        Method & Type of constraints & Model & Exact & Applicable to FOL & Max \# vars. \\
        \midrule
        CGVAE~\cite{liu2018constrained} & Atom valency & VAE & \cmark & \xmark & \\
        CCGVAE~\cite{rigoni2020conditional} & Atom valency & VAE & \cmark & \xmark & \\
        ConStruct~\cite{cg_madeira2024generative} & Graph minor & Diffusion & \cmark & \xmark & \\
        Lifted algorithms~\cite{wang2024lifted} & FOL & WFOMS & \cmark & \cmark & 2 \\
        Ours & FOL & Diffusion & \xmark & \cmark & Any \\
        \bottomrule
        
    \end{tabular}%
    \label{tab:sota}
\end{table*}
\end{mycomment}
Constrained Graph Generation (CGG) can be based on deep learning or, in the specific case of FOL rules, on Weighted First Order Model Sampling (WFOMS). 

For deep-learning-based CGG, the application of constraints to deep graph generative models remains in its infancy, as very few works tackle the problem, or are limited to a specific class of constraints.
Constrained Graph Variational Auto Encoders (CGVAE)~\cite{liu2018constrained} and Conditional CGVAE~\cite{rigoni2020conditional} enforce valency constraints in the molecular domain, closely related to the out-degree of nodes. The model's generative process is autoregressive, allowing at each step to disable linking with atoms that have filled up their valency. ConStruct~\cite{cg_madeira2024generative} is a diffusion model where the whole diffusion and denoising sequences remain in the constrained space, making the samples valid by design. However, the classes of constrains considered are limited to minor-hereditary properties that, if true for a graph, are also true for its minors. 

The generation of graphs can also be seen as the sampling of variables assignments satisfying the set of FOL rules. This has been explored in Weighted First Order Model Sampling (WFOMS)~\cite{wang2024lifted}, closely related to Weighted First Order Model Counting (WFOMC)~\cite{van2011lifted}. Lifted (i.e., polynomial) sampling algorithms exist up to the $C^2$ fragment of FOL~\cite{wang2024lifted}, that is, allowing sentences with universal and counting existential quantifiers $\forall,\exists_{\geq k},\exists_{\leq k},\exists_{=k}$ with up to two variables per sentence. Although exact in its sampling, lifted WFOMS's limitation to two variables could render it ineffective for certain applications, such as ontologies with over 3 variables~\cite{thanapalasingam2023intelligraphs}.

Regarding Knowledge Graph (KG) generation, where FOL ontologies are usually defined, there are two prominent KG completion methods: (1) entity embedding and probabilistic link prediction~\cite{peng2023knowledge}; (2) using Large Language Models (LLM)~\cite{kg_yao2025llm_kg}. The former is designed to predict missing links by computing the likelihood of new triples from the trained embeddings and not to generate new entities. The latter treats triples sets as a textual sequence, and new triples are generated as new tokens.
However, knowledge graph embedding assumes link independence during sampling, limiting the expressiveness of the resulting graph distributions; at the same time, treating graphs as text negates the inductive biases that facilitate scalability to larger structures. Thanks to the availability of KG data and their ontologies, we will evaluate our methodology on KG generation with FOL constraints.
\section{Preliminaries}

We present a formal definition of labeled graphs. A labeled graph $G$ is a tuple $G=(\verts,\edges, \Lambda_v,\Lambda_e)$ where $\verts$ is the set of nodes, $\edges\subseteq\verts\times\verts$ is the set of edges, \mbox{$\Lambda_v:\verts\to\labels_v$} and $\Lambda_e:\edges\to\labels_e$ are the node and edge labeling functions, where $\labels_v$ and $\labels_e$ are the node and edge labels, respectively. Its tensor representation $\fG=(\fX, \fE)$ is a more generation-friendly format, where $\fX\in\R^{n\times D_v}$ is the node tensor, \mbox{$\fE\in\R^{n\times n\times D_e}$} is the edge tensor, $n$ is the number of nodes and $D_v=|\labels_v|$ and $D_e=|\labels_e|+1$, comprising the absence of an edge as one of the $D_e$ classes.

\subsection{Diffusion Models for Graphs}\label{sec:back/diffusion}
Diffusion models~\cite{diff_ho2020diffusion,diff_austin2021discdiff} are a class of generative models where the objective is to learn an approximation $p_\theta(G)$ with parameters $\theta$ of an unknown data distribution $q(G)$. DDPMs achieve this as a Markov Chain Monte Carlo (MCMC) process, reaching convergence to the learned distribution after a certain amount of Markov transitions.
Diffusion models for graphs~\cite{gg_vignac2023digress} jointly generate nodes, edges, and their labels. Two discrete diffusion processes~\cite{diff_austin2021discdiff} randomly flip node and edge labels in $\fX$ and $\fE$, potentially removing or adding connectivity, proceeding from time $t=0$ to $t=T$. The forward and trained reverse processes are defined as:
\begin{align}
    q(\fG^{1:T}|\fG^0) &= \prod_{t=1}^T q(\fX^t|\fX^{t-1})q(\fE^t|\fE^{t-1}), \label{eq:forward}\\
    p_\theta(\fG^{0:T}) &= p(\fG^T)\prod_{t=1}^T p_\theta(\fG^{t-1}|\fG^t). \label{eq:reversed}
\end{align}
We adopt the discrete graph diffusion formulation of DiGress~\cite{gg_vignac2023digress}, which corrupts node and edge labels through categorical transitions and learns the reverse transition $p_\theta(\fG^{t-1}|\fG^t)$. Our method leaves training unchanged and modifies this transition only at inference time.

\subsection{First Order Logic}\label{sec:back/fol}

First Order Logic (FOL) deals with the formalization of facts, which can be either true or false. FOL operates on a first order language, described by a \textit{signature}:
\begin{equation}\label{eq:signature}
    \Sigma=(\constants, \functions, \predicates),
\end{equation}
where $\constants$ is a set of \textit{constant} symbols, $\functions$ is a set of \textit{function} symbols, and $\predicates$ is a set of \textit{predicate} symbols. An interpretation $\interp$ over $\Sigma$ maps each constant $c$ to a specific individual $\interp(c)$ in a domain of discourse $\domain$, function symbols $f$ to functions $\interp(f):\domain^n\to\domain$ of arity $n$, and predicate symbols $P$ to relations $\interp(P)\subseteq\domain^n$ of arity $n$.

Formulas $\phi$ are strings composed of \textit{variables} symbols (e.g., $x,y,z$), non-logical symbols contained in the signature $\Sigma$, and logical symbols, comprising \textit{connectives} (conjunction, disjunction, negation) and \textit{quantifiers} (existential and universal). The semantics of the logic defines the meaning of the logical operators. For example, classical FOL considers only truth values $\{0,1\}$ (false and true), while fuzzy logic also considers intermediate values in $[0,1]$. A sentence, i.e., a formula with no unquantified variables, is \textit{satisfied} by an interpretation $\interp$ if applying $\interp$ to symbols in $\Sigma$ evaluates to true given the semantics of the logic.
\section{Problem Definition}\label{sec:challenges}

Generating labeled graphs $G$ with constraints $\phi$ entails two main challenges: 
\begin{enumerate}
    \item To translate any  constraints into the space of graphs, defining relations over nodes, edges and their labels. This requirement allows checking the constraint satisfaction over graphs, and generalize it as a function $\phi(G)$, which is $1$ if the graph $G$ satisfies the constraints, $0$ otherwise. 
    \item To generate with a model $p_\theta(G)$ graphs that satisfy $\phi$.
\end{enumerate}

Given an existing knowledge base in FOL, defining {\em rules over graphs} can be easily done when there exists a 1:1 correspondence between nodes and the individuals in $\domain$ of the signature $\Sigma$.
However, if $\labels_v=\domain$, e.g., the entities in a knowledge graph, this assumption is broken by $\Lambda_v$ not being injective in general. This carries to graph generative models, which do not assume exclusivity of labels across nodes.

For the second challenge, we want to sample labeled graphs from a constrained distribution, hence restricting the support of a trained generative model $p_\theta(G)$ depending on constraints $\phi$, leading to a conditioned distribution $p_\theta(G|\phi)$. Through logic, we can define rules on the existence, the labels and properties of nodes and edges. We focus on FOL formulas as a very general language for expressing constraints in graph generation, and make graph diffusion models applicable in these contexts, e.g., knowledge graph generation~\cite{thanapalasingam2023intelligraphs}.

Let $p_\theta(\phi(G)=1)$ be the marginal probability of generated graphs satisfying $\phi$. In probabilities, we will also use $\phi$ as a shorthand to indicate the event $\{\phi(G)=1\}$.

{\em Constrained graph generation} entails sampling from a conditional distribution on graphs $p_\theta(G|\phi)$ such that \mbox{$\phi(G)=1$} for all the graphs in its support.
The optimal solution $p^*_\theta(G|\phi)$ to the constrained graph generation problem can be trivially found by applying the Bayes theorem:
\begin{equation}\label{eq:cond_prob}
    p^*_\theta(G|\phi)=\frac{p(\phi|G)p_\theta(G)}{\sum_{W}p(\phi|W)p_\theta(W)}=\frac{\phi(G)p_\theta(G)}{\sum_{W}\phi(W)p_\theta(W)}.
\end{equation}
Computing Eq.~\eqref{eq:cond_prob} is infeasible, as it amounts to solving Weighted First Order Model Counting (WFOMC)~\cite{van2011lifted}, which has computational complexity $\sharp P$~\cite{valiant1979complexity}. We use the exact conditional distribution as an ideal target. Our guided transition is a heuristic approximation designed to increase its support mass on satisfying graphs, so it does not guarantee exact conditional sampling.
\section{Expressing FOL Formulas on Labeled Graphs}\label{sec:signature}
 
To evaluate logic formulas on graphs, we define a signature, starting from Eq.~\ref{eq:signature}, translating labeled nodes and edges into predicates:
\begin{equation}
    \Sigma(G)=(\verts, \emptyset, \labels_v\cup\labels_e\cup\predicates\cup\relations), \label{eq:signature_graph}
\end{equation}
where the constants are all the vertices, no functions are considered, and the predicates are $\labels_v$ and $\labels_e$ for node and edge labels, and $\predicates$ and $\relations$ for additional predicates over nodes and edges. Predicates in $\labels_v$ and $\predicates$ are unary, while those in $\labels_e$ and $\relations$ are binary. The Herbrand interpretation~\cite{detlovs2021introduction} $\interp$ is defined as follows:
\begin{align}
    \interp(v) &= v & \forall v\in\verts\\
    \interp(l_v) &= \{v \mid v\in\verts, \Lambda_v(v)=l_v\} & \forall l_v\in\labels_v\\
    \interp(l_e) &= \{e \mid e\in\edges, \Lambda_e(e)=l_e\} & \forall l_e\in\labels_e\\
    \interp(P) &\subseteq \verts &\forall P\in\predicates\\
    \interp(R) &\subseteq \edges &\forall R\in\relations
\end{align}
With this signature and interpretation, it is now possible to evaluate sentences $\phi$ in FOL over labeled graphs. We denote the formula given a graph $G$ as $\phi(G)$, which, when evaluated in FOL, returns values in $\{0,1\}$.
To make the formulation consistent with the labeling functions of graphs, we will assume that the following exclusivity axioms are always present in the set of constraints $\phi$:
\begin{gather}
    \forall x\bigvee_{l_v\in\labels_v} \left(l_v(x) \land\neg\left(\bigvee_{l'_v\in\labels_v\setminus\{l_v\}}l'_v(x)\right)\right), \\
    \forall x,y\bigvee_{l_e\in\labels_e} \left(l_e(x,y) \land\neg\left(\bigvee_{l'_e\in\labels_e\setminus\{l_e\}}l'_e(x,y)\right)\right).
\end{gather}
Informally, these axioms state that each node and edge must have one and only one label.
When unary and binary predicates $\predicates$, $\relations$ are defined on subset of labels, e.g., a node $v$ with label $horror(v)$ being a $genre(v)$, axioms can be added to the set of rules, of the form:
\begin{gather}
    \forall x \left(P(x) \leftrightarrow \bigvee_{l_v\in\labels_P}l_v(x)\right), \\
    \forall x,y \left(R(x,y) \leftrightarrow \bigvee_{l_e\in\labels_R}l_e(x,y)\right),
\end{gather}
where $\labels_P\subseteq\labels_v$ and $\labels_R\subseteq\labels_e$ are the subsets of labels for which the predicates hold.
These axioms are necessary for a consistent formulation, but have no real consequence on the methodology, nor are they applied. Thus, we mention them now and make them implicit for the rest of the paper.
\section{FOL-Constrained Graph Generation}\label{sec:method}

\begin{figure*}[t]
    \centering
    \includegraphics[width=1.0\textwidth]{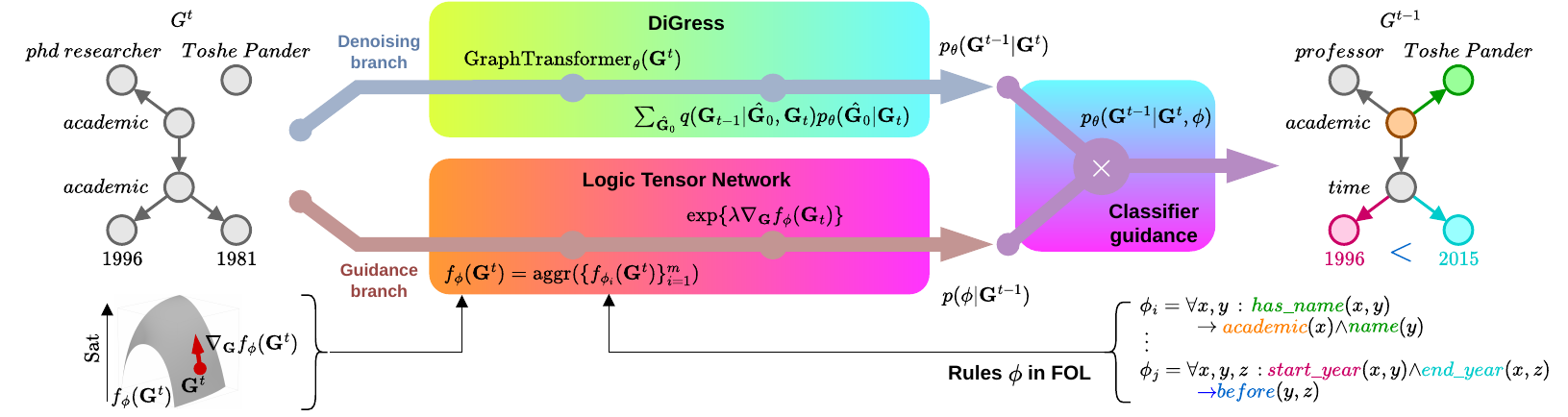}%
    \caption{Our LTN-guidance comprises two branches at each denoising step: (top) the pretrained DiGress model computes the denoising probabilities $p_\theta(\fG^{t-1}|\fG^t)$; (bottom) LTN interprets the given FOL rules $\phi$ as the differentiable function $f_\phi$, over which the gradient on $\fG^t$ is computed, producing an approximation of $p(\phi|\fG^{t-1})$. Classifier guidance (right) combines the two with element-wise product, returning the conditional probabilities $p_\theta(\fG^{t-1}|\fG^t,\phi)$.}
    \label{fig:method_overview}%
\end{figure*}

For FOL constrained graph generation, we leverage classifier guidance to steer a pretrained graph diffusion model towards satisfying a formula $\phi$. Classifier guidance requires a differentiable objective function for gradient computation, so we turn to Logic Tensor Networks (LTN) to interpret $\phi$ in fuzzy logic. The contributions from each sentence in $\phi$ are reweighted through a mechanism based on the Focal Loss. An overview of LTN-guidance can be found in Fig.~\ref{fig:method_overview}.

{\em Classifier guidance}~\cite{dhariwal2021diffusion} is one of the most popular alternative for conditioning diffusion models. Intuitively, the denoising path is steered towards the optimization of a classifier prediction. Although the name suggests the use of a trained classifier, in practice it can be any differentiable function, and may not require training. In our case, the classifier is implemented by the formula satisfaction:
\begin{equation}\label{eq:cls_guid_trans}
    \begin{aligned}
        &p_\theta(\fG^{t-1}|\fG^t,\phi)\propto p(\phi|\fG^{t-1})p_\theta(\fG^{t-1}|\fG^t)\\
        &\ \approx\exp(\lambda \fG^{t-1\top} \nabla_{\fG}\log p(\phi(\fG^t)=1|\fG^t))\ p_\theta(\fG^{t-1}|\fG^t),
    \end{aligned}
\end{equation}
where $p_\theta(\fG^{t-1}|\fG^t)$ is the unconditional model denoising transition from Eq.~\ref{eq:reversed}, and $\lambda$ is the guidance strength, tunable as a hyperparameter. The approximation comes from the first order Taylor approximation of $p(\phi|\fG^{t-1})$ evaluated at $\fG^t$. For a general proof, see~\citet{gg_vignac2023digress}. In classic FOL, $ p(\phi(\fG^t)=1|\fG^t)$ is not differentiable, as it is equivalent to $\phi(G^t)$, taking values only in $\{0,1\}$. Thus, we propose the use of fuzzy logic, and in particular Logic Tensor Network (LTN)~\cite{badreddine2022logic}, to provide differentiable fuzzy interpretations of the formulas.
We implement $\log p_\phi(\phi(\fG^t)=1|\fG^t)$ as a differentiable function $f_\phi(\fG^t)$.

{\em Logic Tensor Network}~\cite{badreddine2022logic} provides differentiable fuzzy semantics for FOL predicates, connectives, and quantifiers. Fuzzy logic allows to specify the semantic $\semantics$ of logical operators to return values in the range $[0,1]$. These functions behave exactly the same as the classical FOL semantics in the case of inputs in $\{0,1\}$, but allow intermediate values, usually representing a degree of truth. In the following, we define LTN applied to a tensor graph $\fG$. We start from the base case of generic unary and binary predicates $P_i$ and $R_j$:
\begin{align}
    \semantics_{\fG}(P_i(v)) &= \sum_{l\in\labels_{P_i}}\fX_{vl}, \\
    \semantics_{\fG}(R_j(v,v')) &= \sum_{l\in\labels_{R_j}}\fE_{vv'l},
\end{align}
where we remark that $\verts$ is both the set of nodes in $G$ and the constants of the signature $\Sigma(G)$.
We continue recursively with the semantics of logical connectives in product logic:
\begin{align}
    \semantics_{\fG}(\phi\land\psi)&= \semantics_{\fG}(\phi)\semantics_{\fG}(\psi), \\
    \semantics_{\fG}(\phi\lor\psi)&= 1-(1-\semantics_{\fG}(\phi))(1-\semantics_{\fG}(\psi)), \\
    \semantics_{\fG}(\neg\phi)&= 1-\semantics_{\fG}(\phi),
\end{align}
and all quantifiers with the following aggregation functions from~\citet{badreddine2022logic}:
\begin{align}
    \semantics_{\fG}(\forall x\phi(x))&= 1-\left(\frac{1}{|\verts|}\sum_{v\in\verts}\Big(1-\semantics_{\fG}(\phi)_v\Big)^p\right)^{\frac{1}{p}},\\
    \semantics_{\fG}(\exists x \phi(x))&= \left(\frac{1}{|\verts|}\sum_{v\in\verts}\Big(\semantics_{\fG}(\phi)_v\Big)^p\right)^{\frac{1}{p}}.
\end{align}
In the case of quantification over 2 variables, it is sufficient to aggregate the p-mean over all pairs $(v_i,v_j)\in\verts\times\verts$.

Some setups~\cite{thanapalasingam2023intelligraphs} may append {\em non-trivial constraints to FOL statements}, whose direct grounding under the standard LTN formulation is computationally impractical. This is the case of edge cardinality constraints, which require to have exactly $k$ edges. This type of constraint can be implemented as a FOL rule for fixed $k$:
\begin{equation}
    \begin{aligned}
        &\phi_{\text{card},k} : \ \exists x_1,y_1,\dots,x_k,y_k\Bigg(\bigwedge_{i=1}^k R(x_i,y_i) \\
        &\ \ \land\bigwedge_{i\neq j}\neg(x_i=x_j\land y_i=y_j) \\
        &\ \ \land\forall x,y\left(R(x,y)\to\bigvee_{i=1}^k (x=x_i\land y=y_i)\right)\Bigg),
    \end{aligned}
\end{equation}
but will require the introduction of $k+1$ pairs of variables, making checks computationally expensive. Thus, we implement a custom and easy to compute differentiable function for cardinality constraints:
\begin{equation}
    \semantics_{\fG}(\phi_{\text{card},k})=\exp\left(-\frac{|\sum_{v,v'\in\verts,l\in\labels_e}\fE_{vv'l}-k|^p}{\sigma}\right),\label{eq:cardinality}
\end{equation}
where $\sigma$ is a positive temperature parameter, and $p$ is the same exponent used for the quantifiers.

With LTN, a sentence $\phi$ can now be translated to a differentiable function $f_{\phi}(\fG)=\semantics_{\fG}(\phi)$. Still, the overall constraint may be a conjunction of multiple sentences $\phi_1,\dots,\phi_m$.
Let $\phi$ be their conjunction.
To each sentence $\phi_i$ corresponds its fuzzy interpretation $f_{\phi_i}$.
To aggregate the rules into a {\em single differentiable function}, we apply an adaptive reweighting method based on the Focal Loss~\cite{lin2017focal}:
\begin{equation}\label{eq:agg_logic}
    f_\phi(\fG)=\sum_{i=1}^m(1-f_{\phi_i}(\fG)) \log(f_{\phi_i}(\fG)).
\end{equation}
When a sentence is satisfied, i.e., $f_{\phi_i}(\fG)\approx 1$, then the contribution of $\log(f_{\phi_i}(\fG))$ disappears, leaving unsatisfied sentences to steer the generative model. To apply LTN-guidance it is enough to replace $\nabla_{\fG}\log p_\phi(y|\fG_t)$ with $\nabla_{\fG}f_\phi(\fG_t)$.
\section{Experiments}\label{sec:exp}

In our experimentation, we consider three research questions:
\textbf{RQ1:} Does LTN-guidance increase the satisfaction of dataset rules?
\textbf{RQ2:} How does LTN-guidance affect the denoising process?
\textbf{RQ3:} Does LTN-guidance allow the addition of new rules to either restrict the support or shift it out of distribution?

\subsection{Benchmark}~\label{sec:exp/dataset}
We use the IntelliGraphs benchmark~\cite{thanapalasingam2023intelligraphs} to evaluate FOL-guided graph generation. IntelliGraphs contains five Knowledge Graphs (KGs) datasets designed to test different forms of semantic validity. \texttt{syn-paths} tests directed path structure; \texttt{syn-types} tests compatibility between entity and relation types together with a fixed-cardinality constraint; and \texttt{syn-tipr} instantiates a time-indexed person-role ontology pattern. The two real-world datasets contain Wikidata-derived subgraphs: \texttt{wd-movies} describes a movie through actors, directors, and genres, whereas \texttt{wd-articles} describes scholarly articles and ordered authorship information. The latter is the most structurally complex because valid graphs must represent consecutive author positions without gaps. This same benchmark was used in~\citet{wang2025provcreator}, and the amount of available data is substantially larger than existing generic graph datasets~\cite{gg_huang2023cdgs}.
Graphs in the datasets are in Resource Description Framework (RDF)~\cite{cyganiak2014rdf} format, i.e., triples $\triples=\{(s,r,o):s,o\in\domain,r\in\relations\}$ where $s$ is the subject entity, $r$ the edge predicate or relation, $o$ the object entity, and $\domain$ and $\relations$ are the entity and relation sets, respectively.
To make them compatible with our generative model, we converted\footnote{We identified some irregularities in the datasets, such as the representation of triples $\triples$ not as sets but as lists, which allows for duplicates. For example, the \texttt{syn-types} dataset contains KGs with 10 triples each, but graph conversion shows that roughly the $76.25\%$ of the sets actually have 10 triples, while the remaining $23.75\%$ have less than 10 triples.} all of them to graphs $G$.
A detailed introduction to knowledge graphs, and how to formally convert them to labeled graphs can be found in Appendix C of the supplementary material.

The IntelliGraphs benchmark considers the percentage of graphs that are valid, novel, valid and novel, or empty. {\em Validity} is one of the main indicators of the efficiency for generative models, as it directly influences the number of resamples to reach a target number of valid samples. The percentage of valid graphs  (i.e., graphs for which $\phi(G)=1$) in a batch of samples $\mathcal{S}$ is:
\begin{equation}\label{eq:valid}
    V(\phi,\mathcal{S})=\frac{1}{|\mathcal{S}|}\sum_{G\in\mathcal{S}}\phi(G).
\end{equation}
In IntelliGraphs, the validity computation is implemented with a checker that verifies each rule in the ontology\footnote{We would like to point out that we found a discrepancy between the rules implemented in the checker and those presented in the paper. For example, the \texttt{wd-movies} dataset contains a rule stating that all KGs should have at least a genre. Yet, it is not present in the checker, and many KGs of the dataset do not contain any genre. Thus, we use the implemented checker in our methodology.}.
Two KGs are considered equal if their set of triples (subject, predicate, object) are the same. Given this notion, {\em Novelty} is computed on a batch $\mathcal{S}$ and training set $\mathcal{D}_\text{tr}$ as the percentage of sampled graphs not present in the training set:
\begin{equation}\label{eq:novel}
    N(\mathcal{S}) = \frac{|\mathcal{S}\setminus\mathcal{D}_\text{tr}|}{|\mathcal{S}|}.
\end{equation}
The benchmark also considers the percentage of novel graphs $\mathcal{S}\setminus\mathcal{D}_\text{train}$ that are also valid (\textit{Valid and Novel}) and the percentage of generated graphs in $\mathcal{S}$ with no edges, formally graphs where $\triples=\emptyset$ (\textit{Empty Graphs}).

We included additional metrics from the graph generation literature~\cite{thompson2022evaluation} to evaluate how well the generative model captures the distribution of the data. These are based on the Maximum Mean Discrepancy (MMD) between two sets of graphs, computing pairwise kernel functions $k(\cdot,\cdot)$ on features extracted from graphs using a feature extractor $f$. MMD is defined as:
\begin{equation}
    \begin{aligned}
        &\text{MMD}^2(p,q) = \mathbb{E}_{G,G'\sim p}[k_f(G,G')] +  \\
        & \ \ + \mathbb{E}_{G,G'\sim q}[k_f(G,G')] - 2\mathbb{E}_{G\sim p,G'\sim q}[k_f(G,G')].
    \end{aligned}
\end{equation}
As features, we consider the node degrees (\textit{Degree}), the clustering coefficients (\textit{Clustering}), and the eigenvalues of the graph Laplacian (\textit{Spectre}). As kernel, we use the radial basis function.

\subsection{Implementation Details}\label{sec:exp/impl}
We apply LTN-guidance to DiGress~\cite{gg_vignac2023digress}, a state-of-the-art graph diffusion model. Because DiGress was originally implemented for undirected graphs, we reimplemented parts of the model and Graph Transformer~\cite{dwivedi2020graph_transf} to avoid symmetries in the adjacency matrix, e.g., removing matrix symmetrization operations $\adj\adj^\top$. This allows DiGress to be used for the generation of directed graphs.
We compare DiGress against ComplEx~\cite{trouillon2016complex}, the best KG embedding baseline of IntelliGraphs, and ProvCreator~\cite{wang2025provcreator}, an LLM-based model. Specialized constrained generation methods are not applicable to this setup, e.g., ConStruct~\cite{cg_madeira2024generative} only handles minor-hereditary properties, which do not have any overlap with the benchmark. We note that these baselines directly use the triples format, while we use their labeled graph transformation. Furthermore, ComplEx receives the correct entity set, which is not true for other methods. For our guided models, we used standard LTN values for $p$, while the guidance strength was split into $\lambda_x$ for nodes and $\lambda_e$ for edges gradients, and their validity-optimizing value per dataset was found through grid search in the range $[10^{-5},1]$. All experiments were conducted on a single A40 GPU. More details on the hyperparameter search and selected values can be found in Appendix A of the supplementary material.
\subsection{Results and Discussion}\label{sec:exp/results}
\begin{table*}[t]
    \fontsize{7}{9}\selectfont
    \setlength{\tabcolsep}{3pt} 
    \centering
    \caption{Results on the IntelliGraphs benchmark. Because the setup is the same, the reported values for ComplEx and ProvCreator are taken from~\cite{thanapalasingam2023intelligraphs} and~\cite{wang2025provcreator}, respectively. For our method, we report \meanstd{mean}{std} across three models trained with random seeds from 0 to 2, sampling 1,000 graphs per dataset, except for 200 graphs on \texttt{wd-articles}.}
    \begin{tabular}{@{}cc*{8}{c}@{}}
        \toprule
        Dataset & Model & \%Valid$\uparrow$ & \%Nov. \& Val.$\uparrow$ & \%Novel$\uparrow$ & \%Empty$\downarrow$
        & Degree$\downarrow$ & Clustering$\downarrow$ & Spectre$\downarrow$ \\
        \midrule
        
        \multirow{4}{*}{\texttt{syn-paths}}
        & ComplEx (w/ $p(S|E)$ only) & 10.10 & 10.10 & 95.58 & 4.42 & $-$ & $-$ & $-$\\
        & ProvCreator & 93.14 & 93.14 & 100.00 & 0.00 & $-$ & $-$ & $-$\\
        & DiGress (no guid) & \meanstd{99.43}{0.31} & \meanstd{99.40}{0.30} & \meanstd{99.97}{0.06} & \meanstd{0.00}{0.00} & \meanstd{1.6e-5}{9.17e-6} & \meanstd{1.33e-6}{1.15e-6} & \meanstd{1.6e-5}{9.17e-6}\\
        & DiGress (w/ guid) & \meanstd{99.40}{0.44} & \meanstd{99.37}{0.42} & \meanstd{99.97}{0.06} & \meanstd{0.00}{0.00} & \meanstd{2.13e-5}{1.81e-5} & \meanstd{1.33e-6}{1.15e-6} & \meanstd{2.13e-5}{1.81e-5}\\
        \midrule
        
        \multirow{4}{*}{\texttt{syn-tipr}}
        & ComplEx (w/ $p(S|E)$ only) & 0.00 & 0.00 & 99.64 & 0.36 & $-$ & $-$ & $-$\\
        & ProvCreator & 0.00 & 0.00 & 100.00 & 0.00 & $-$ & $-$ & $-$\\
        & DiGress (no guid) & \meanstd{99.70}{0.36} & \meanstd{95.87}{1.07} & \meanstd{96.17}{0.71} & \meanstd{0.00}{0.00} & \meanstd{3e-5}{1.74e-5} & \meanstd{6.67e-7}{1.15e-6} & \meanstd{3e-5}{1.74e-5} \\
        & DiGress (w/ guid) & \meanstd{99.70}{0.36} & \meanstd{95.87}{1.07} & \meanstd{96.17}{0.71} & \meanstd{0.00}{0.00} & \meanstd{3e-5}{1.74e-5} & \meanstd{6.67e-7}{1.15e-6} & \meanstd{3e-5}{1.74e-5} \\
        \midrule

        \multirow{4}{*}{\texttt{syn-types}}
        & ComplEx (w/ $p(S|E)$ only) & 1.01 & 1.01 & 94.17 & 5.83 & $-$ & $-$ & $-$\\
        & ProvCreator & 86.16 & 86.16 & 100.00 & 0.00 & $-$ & $-$ & $-$\\
        & DiGress (no guid) & \meanstd{74.97}{0.85} & \meanstd{74.97}{0.85} & \meanstd{100.00}{0.00} & \meanstd{0.00}{0.00} & \meanstd{1.61e-3}{1.7e-4} & \meanstd{4.19e-4}{5.85e-5} & \meanstd{4.51e-3}{1.52e-4} \\
        & DiGress (w/ guid) & \meanstd{95.70}{1.00} & \meanstd{95.70}{1.00} & \meanstd{100.00}{0.00} & \meanstd{0.00}{0.00} & \meanstd{0.01}{1.12e-3} & \meanstd{0.02}{3.56e-3} & \meanstd{0.01}{4.7e-4} \\
        \midrule

        \multirow{4}{*}{\texttt{wd-movies}}
        & ComplEx (w/ $p(S|E)$ only) & 0.41 & 0.41 & 93.04 & 6.96 & $-$ & $-$ & $-$\\
        & ProvCreator & 94.26 & 94.26 & 100.00 & 0.00 & $-$ & $-$ & $-$\\
        & DiGress (no guid) & \meanstd{93.00}{0.95} & \meanstd{87.63}{1.69} & \meanstd{94.63}{2.40} & \meanstd{0.00}{0.00} & \meanstd{0.05}{0.01} & \meanstd{0.00}{0.00} & \meanstd{0.08}{0.01} \\
        & DiGress (w/ guid) & \meanstd{97.37}{0.40} & \meanstd{91.40}{2.75} & \meanstd{94.03}{2.35} & \meanstd{0.00}{0.00} & \meanstd{0.05}{0.01} & \meanstd{0.00}{0.00} & \meanstd{0.07}{0.01} \\
        \midrule

         \multirow{4}{*}{\texttt{wd-articles}}
         & ComplEx (w/ $p(S|E)$ only) & 0.00 & 0.00 & 100.00 & 0.00 & $-$ & $-$ & $-$\\
        & ProvCreator & 0.00 & 0.00 & 100.00 & 0.00 & $-$ & $-$ & $-$\\
        & DiGress (no guid) & \meanstd{64.50}{7.00} & \meanstd{64.50}{7.00} & \meanstd{100.00}{0.00} & \meanstd{0.00}{0.00} & \meanstd{0.16}{0.02} & \meanstd{0.02}{0.03} & \meanstd{0.14}{0.01} \\
        & DiGress (w/ guid) & \meanstd{69.50}{3.97} & \meanstd{69.50}{3.97} & \meanstd{100.00}{0.00} & \meanstd{0.00}{0.00} & \meanstd{0.15}{0.01} & \meanstd{0.01}{0.01} & \meanstd{0.14}{0.01} \\
        \bottomrule
        
    \end{tabular}%
    \label{tab:kg_results}
\end{table*}

\paragraph{(RQ1) Improving validity on dataset.} We use LTN-guidance to integrate the rules used to check the validity. Our technique increases validity where the unconditional model violates constraints, while preserving near-perfect validity on the other datasets. The results are presented in Table~\ref{tab:kg_results} where, although some metrics and datasets might be redundant, we chose to keep them for the completeness of the benchmark. It can be seen that DiGress, even without guidance, achieves very good results in all datasets, producing practically perfect results on \texttt{syn-paths} and \texttt{syn-tipr}. In \texttt{syn-types} and \texttt{wd-movies}, we observe that unconditional DiGress is outperformed by the ProvCreator baseline. For the former, we find that the source of error is the cardinality constraint of the dataset. In fact, the validity reached by DiGress is very close to the percentage of triples set of length 10 without duplicates in the dataset.
The inclusion of the cardinality constraint in Eq.~\ref{eq:cardinality} improves its satisfaction, leading to the very high validity of $95.70\%$. In \texttt{wd-movies}, LTN-guidance also improves results over vanilla DiGress with an average validity of $97.37\%$, outperforming ProvCreator. Still, classifier guidance did not show substantial improvements on \texttt{wd-articles}, the most challenging dataset. In particular, the most unsatisfied constraint entails having no gaps in the paper author positions. Near the final denoising steps, the unconditional transition becomes sharply concentrated, and the chosen guidance strength is insufficient to change the model's predictions. Increasing the weights was insufficient, so a more complex weighting scheme is needed. Regarding distributional metrics, we observe that guidance does not substantially skew the distribution compared to the unconditional models. This means that the technique does not sacrifice sample quality for validity.

\begin{figure}[t]
    \centering
    \begin{subfigure}{\columnwidth}
        \includegraphics[width=\textwidth]{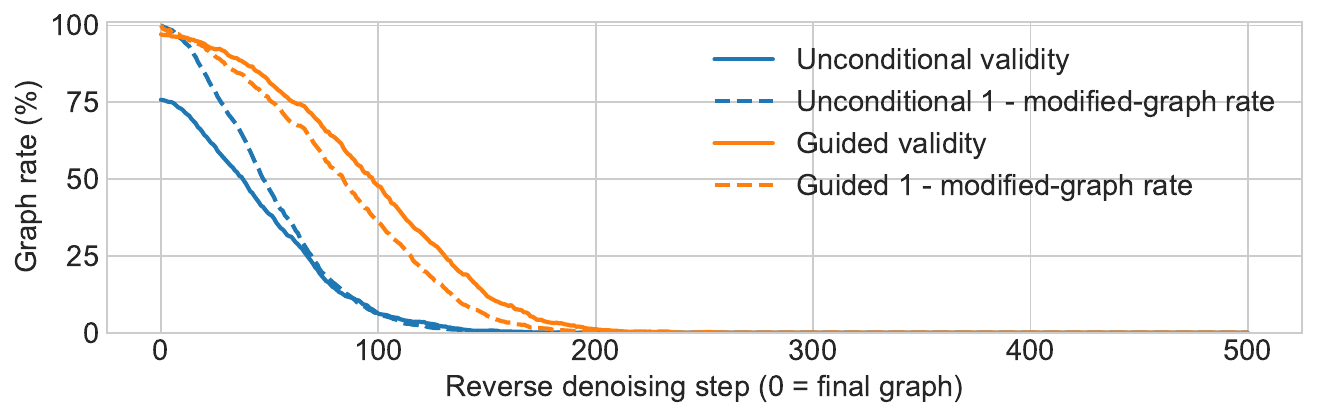}
        \caption{Average validity and $1-R(\mathcal{S}_0,\mathcal{S}_t)$}%
        \label{fig:validity_diff}
    \end{subfigure}
    \begin{subfigure}{\columnwidth}
        \includegraphics[width=\textwidth]{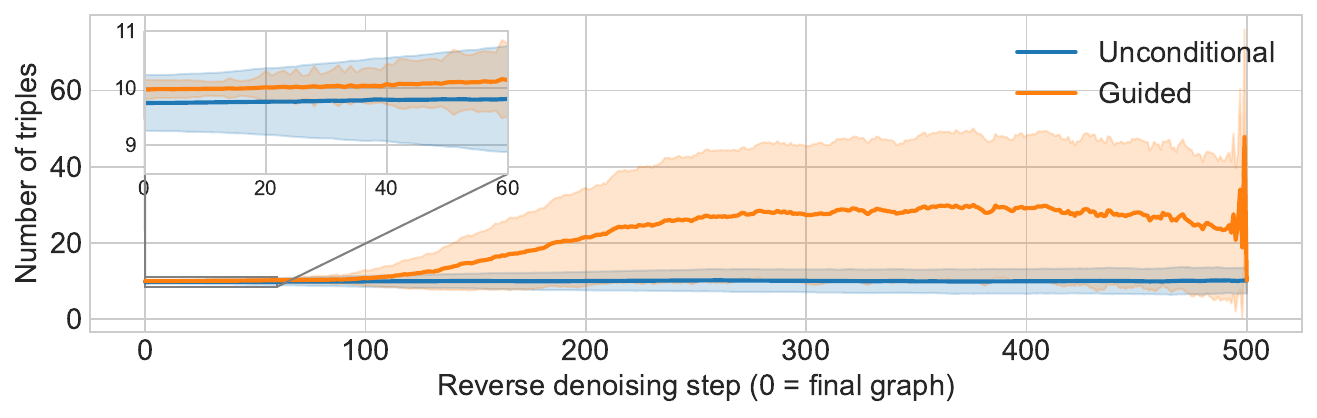}
        \caption{Average number of triples $|\triples|$ (with zoom-in). Target is $|\triples|=10$. Shaded area represents the standard deviation.}%
        \label{fig:num_triples}
    \end{subfigure}
    \caption{Comparison of unconditional and guided denoising processes in the \texttt{syn-types} dataset across 3 seeds.
    }
    \label{fig:validity_vs_modif}
\end{figure}

\paragraph{(RQ2) How LTN-guidance acts during denoising.} We generated a batch of $N=1000$ graphs from the model trained on the \texttt{syn-types} dataset, collecting the trajectories, i.e., the intermediate graphs during denoising, in batches $\mathcal{S}_t=\{G_{t,1},\dots,G_{t,N}\}$. Then, we plotted the validity of all $\mathcal{S}_t$ along time, together with the rate of modification of graphs $G_t$ with respect to the final $G_0$, computed as:
\begin{equation}
    R(\mathcal{S}_0,\mathcal{S}_t)=\frac{1}{|\mathcal{S}|}\sum_{i=1}^N\m1_{>0}\left(\operatorname{GED}(G_{0,i},G_{t,i})\right),
\end{equation}
where $\operatorname{GED}$ is the Graph Edit Distance with unit costs, and $\m1_{>0}$ is the indicator function for the positive set. The plots obtained by gathering $\mathcal{S}_t$ can be found in Fig.~\ref{fig:validity_vs_modif}. We remark that, because the graphs in \texttt{syn-types} converted from triples do not all contain exactly 10 edges, the validity is not high at time $t=0$. A stark difference can be seen with LTN-guidance, where validity reaches almost $100\%$. A key observation is that validity starts to increase many timesteps earlier compared to denoising without guidance, followed by the inverse modification rate, meaning that the final graphs are achieved much earlier. This is further confirmed by the plot in Fig.~\ref{fig:num_triples}, showing the trends in the number of triples $|\triples|$ in the graphs throughout the denoising process. The target is $|\triples|=10$, and the guided denoising overshoots this value, but finally converges to it with low standard deviation, while the unconditional denoising terminates at a $<10$ value.

\begin{figure}[t]
    \centering
    \begin{subfigure}[t]{\columnwidth}
        \centering
        \includegraphics[width=1.\linewidth]{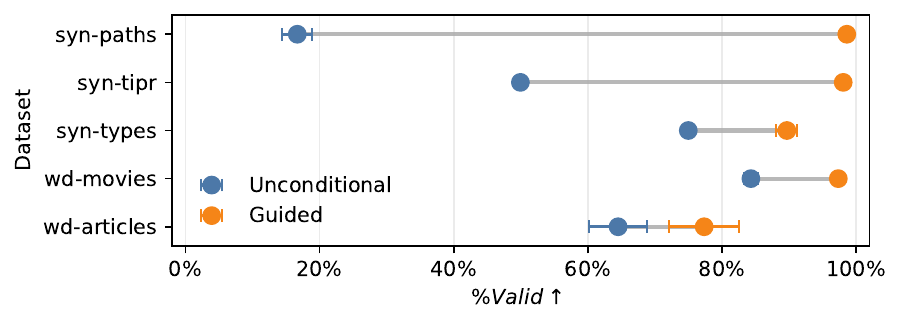}
        \caption{Support restriction rules}
    \end{subfigure}
    \begin{subfigure}[t]{\columnwidth}
        \centering
        \includegraphics[width=1.\linewidth]{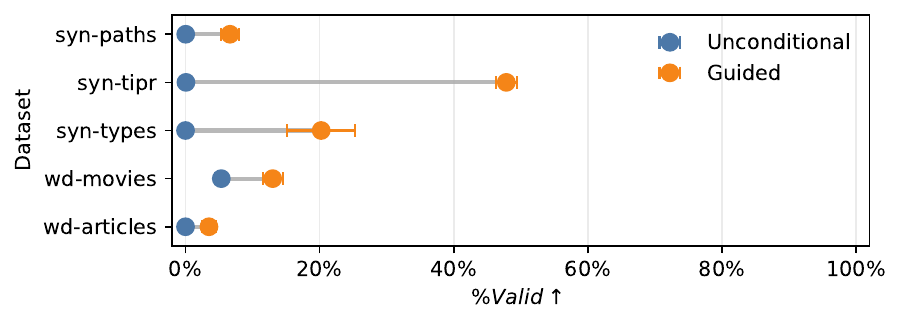}
        \caption{Support shift rules}
    \end{subfigure}
    \caption{Validity comparison of unconditional and guided models when introducing new compatible (restriction) or conflicting (shift) rules.}
    \label{fig:support}
\end{figure}

\paragraph{(RQ3) Plugging new rules.} We evaluate the capability of LTN-guidance to enforce rules not considered in the datasets in two setups: (1) support restriction, where the new rules are compatible with the current knowledge, providing an additional control signal; (2) support shift, where we introduce new conflicting rules, removing broken ones from the data rules. For example, requiring a genre for each KG in \texttt{wd-movies} constitutes a restriction, while changing the number of triples for \texttt{syn-types} to 12 is a shift, as no in-distribution KG satisfies this condition. Details on added rules can be found in Appendix B of the supplementary material. We want to remark that setup 2 is very challenging, corresponding to out-of-distribution generation, which is non-trivial for generative models. Results are shown in Fig.~\ref{fig:support}. We observe that for the support restriction task, LTN-guidance is able to improve in all cases the validity with respect to the new rules, reaching almost 100\% validity in all cases. This means that in-distribution conditioning is possible with LTN-guidance. As mentioned previously, the support shift task is more challenging, but LTN-guidance introduces a non-zero likelihood of generating graphs satisfying the rules. Thus, the guidance signal produced by LTN-guidance may be used to produce out-of-distribution samples.

\subsection{Limitations}
LTN-guidance uses a differentiable relaxation of the logical rules, which does not guarantee perfect validity. Another limitation is the requirement for a thorough search for the best guidance weight $\lambda$ per dataset. This work tackles the generation of graphs with FOL constraints, on which we found a limited body of competing works and data to compare with. Thus, a broader empirical picture is left for future work.
\section{Conclusion}

We presented how to steer graph denoising diffusion models with FOL rules through LTN-guidance. We first defined a graph signature and consistency axioms under which FOL rules can be written; second, we proposed to interpret the rules as differentiable objective functions through LTN; finally, classifier guidance steers the denoising process with the obtained function. Thus, a pre-trained graph diffusion model can be conditioned to maximize the satisfaction of a set of FOL rules, e.g., the ontology for knowledge graphs. This exact setup has been used to evaluate our technique, showing improved validity compared to the unconditional model. We also studied the behavior during the denoising process, observing that FOL classifier guidance acts aggressively, causing samples to reach their final discrete structure earlier in denoising time. In the future, several diverse graph setups might benefit from its applications. For instance, in counterfactual generation for Graph Neural Networks (GNN)~\cite{chen2026generating} one might use LTN-guidance to have control over the counterfactual distribution at inference time. In conditioned molecular generation and retrosynthesis~\cite{qiao2025advancing}, certain elements may be prohibited, or commercially available starting materials might be required.

\section*{Acknowledgments}
We acknowledge the financial support of (i) the PNRR project FAIR — Future AI Research (PE00000013), under the NRRP MUR program funded by the NextGenerationEU; (ii) the Italian Ministry of University and Research (MUR) under the PRIN program --- Progetti di Rilevante Interesse Nazionale --- PRIN 2022 (Secretary-General's Decree No. 1401 of 18/09/2024) --- CUP C53C24000770006, project title “DEEP-GRAPH: Design and Theory of Deep Graph Learning”.

\bibliography{refs}

\clearpage
\def\ARXIVCOMBINED{1}
\ifdefined\ARXIVCOMBINED\else
\documentclass[letterpaper]{article} 
\usepackage[submission]{aaai2027}  
\usepackage[hyphens]{url}  
\usepackage{graphicx} 
\urlstyle{rm} 
\def\UrlFont{\rm}  
\usepackage{natbib}  
\usepackage{caption} 
\frenchspacing  
%
\usepackage{algorithm}
\usepackage{algorithmic}

%
\usepackage{newfloat}
\usepackage{listings}
\DeclareCaptionStyle{ruled}{labelfont=normalfont,labelsep=colon,strut=off} 
\lstset{%
	basicstyle={\footnotesize\ttfamily},
	numbers=left,numberstyle=\footnotesize,xleftmargin=2em,
	aboveskip=0pt,belowskip=0pt,%
	showstringspaces=false,tabsize=2,breaklines=true}
\floatstyle{ruled}
\newfloat{listing}{tb}{lst}{}
\floatname{listing}{Listing}

%
\usepackage{booktabs}

%
\pdfinfo{
/TemplateVersion (2027.1)
}

\setcounter{secnumdepth}{2} 

%


\title{Supplementary material for ``Logic-Guided Graph Denoising Diffusion Models''}
\author{
    Written by AAAI Press Staff\textsuperscript{\rm 1}\thanks{With help from the AAAI Publications Committee.}\\
    AAAI Style Contributions by Peter Patel Schneider,
    Sunil Issar,\\
    J. Scott Penberthy,
    George Ferguson,
    Hans Guesgen,
    Francisco Cruz\equalcontrib\corresponding,
    Marc Pujol-Gonzalez\equalcontrib\corresponding
}
\affiliations{
    \textsuperscript{\rm 1}Association for the Advancement of Artificial Intelligence\\


    1101 Pennsylvania Ave, NW Suite 300\\
    Washington, DC 20004 USA\\
    proceedings-questions@aaai.org
%
}

\usepackage{amsmath}
\usepackage{amsfonts}
\usepackage{booktabs}
\usepackage{multirow}
\usepackage{makecell}
\usepackage{pifont}
\usepackage{subcaption}

\begin{document}

\maketitle
\fi

\appendix

\section{Hyperparameters and Implementation Details}
\begin{figure*}[p]
    \centering
    \begin{subfigure}[t]{0.4\textwidth}
        \centering
        \includegraphics[width=\textwidth]{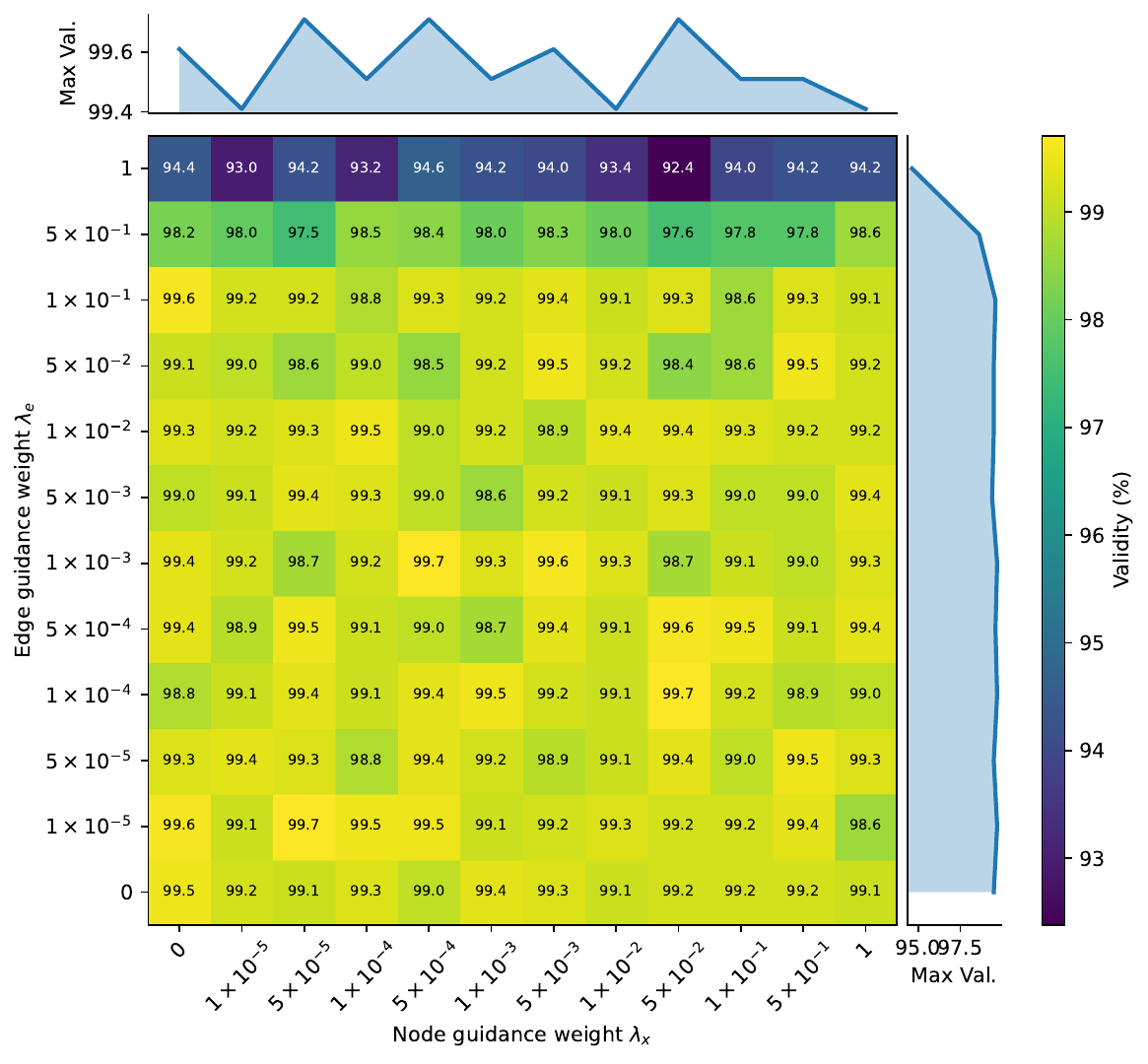}
        \caption{syn-paths}
    \end{subfigure}
    \begin{subfigure}[t]{0.4\textwidth}
        \centering
        \includegraphics[width=\textwidth]{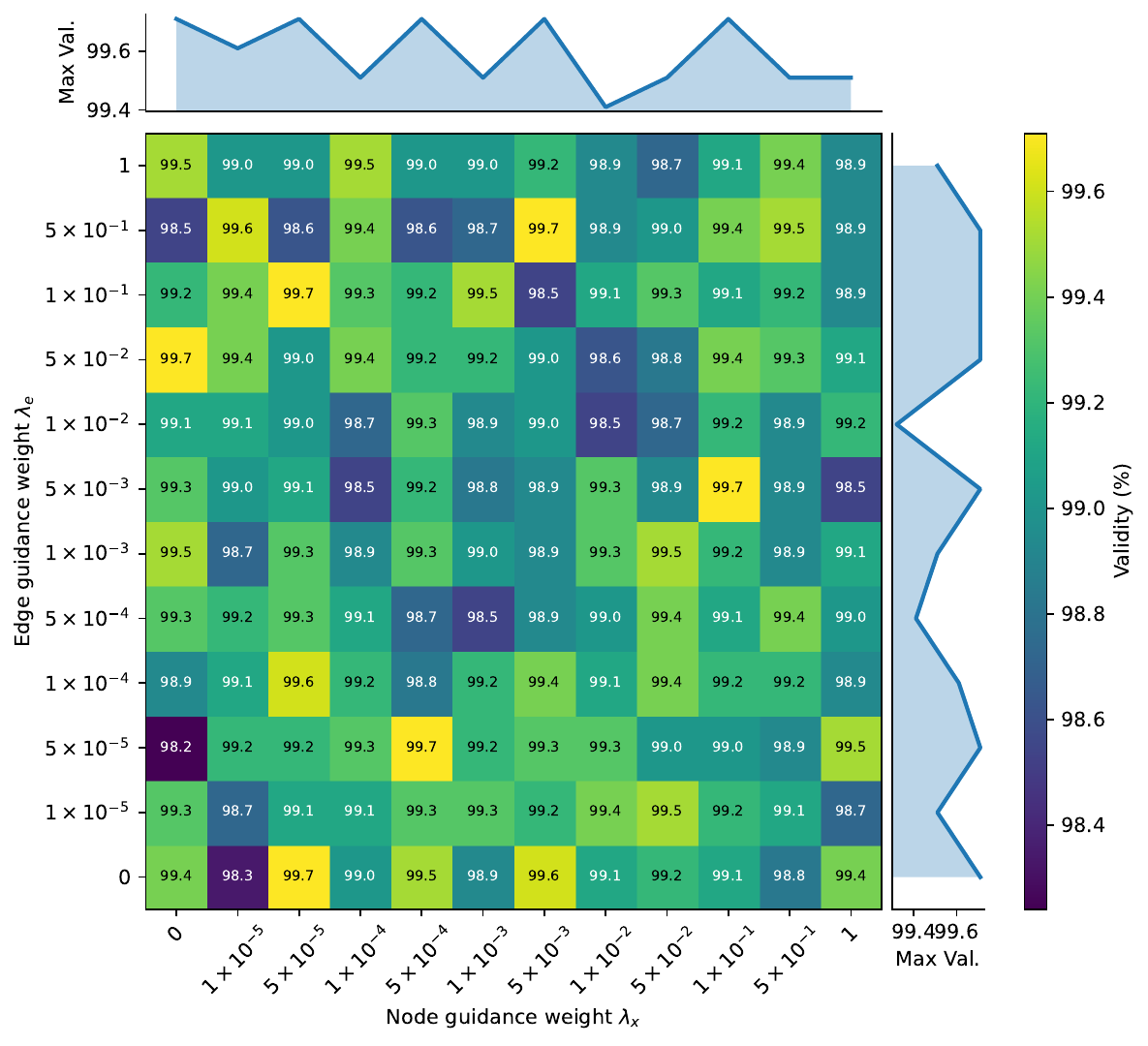}
        \caption{syn-tipr}
    \end{subfigure}
    \begin{subfigure}[t]{0.4\textwidth}
        \centering
        \includegraphics[width=\textwidth]{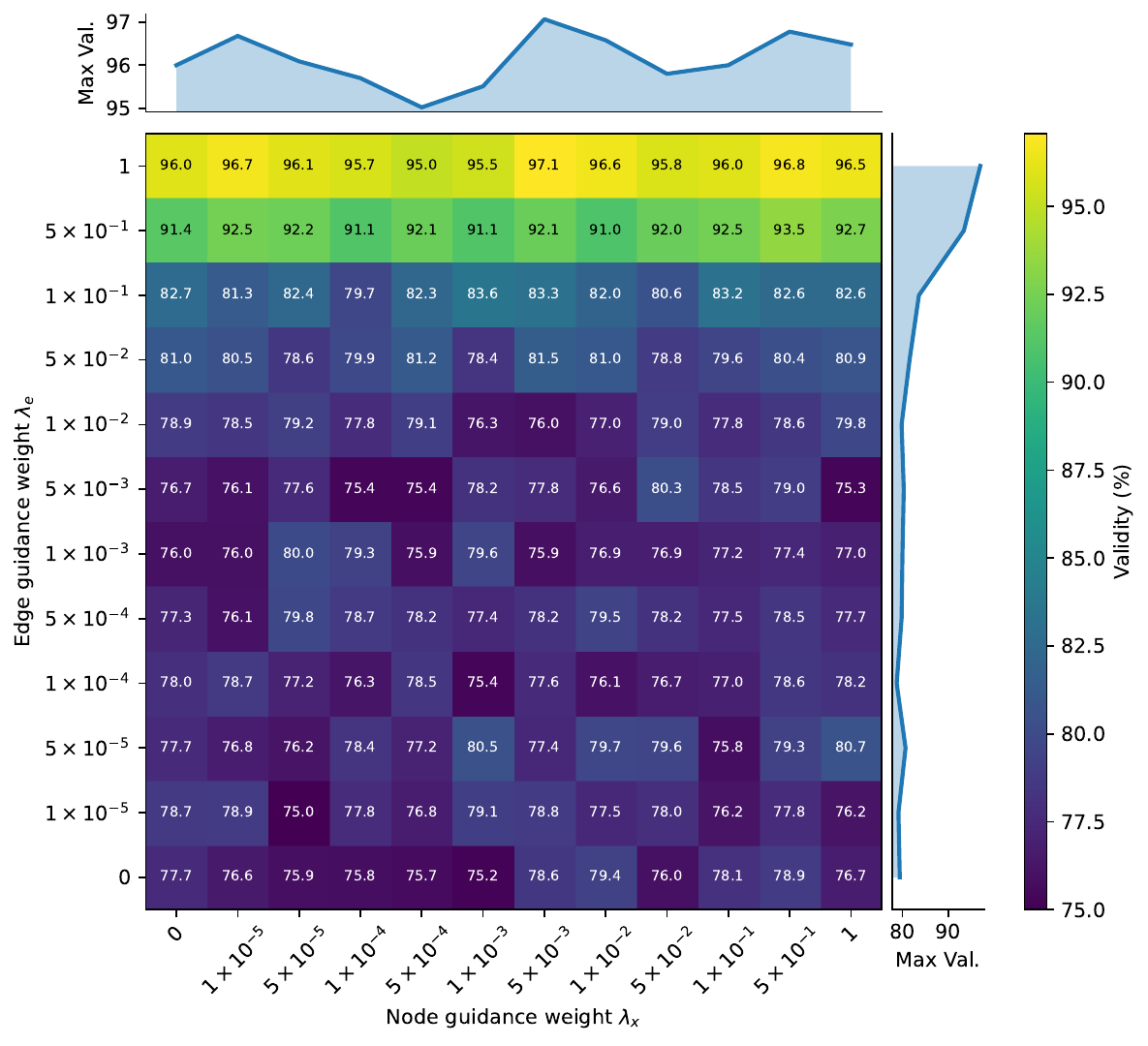}
        \caption{syn-types}
    \end{subfigure}
    \begin{subfigure}[t]{0.4\textwidth}
        \centering
        \includegraphics[width=\textwidth]{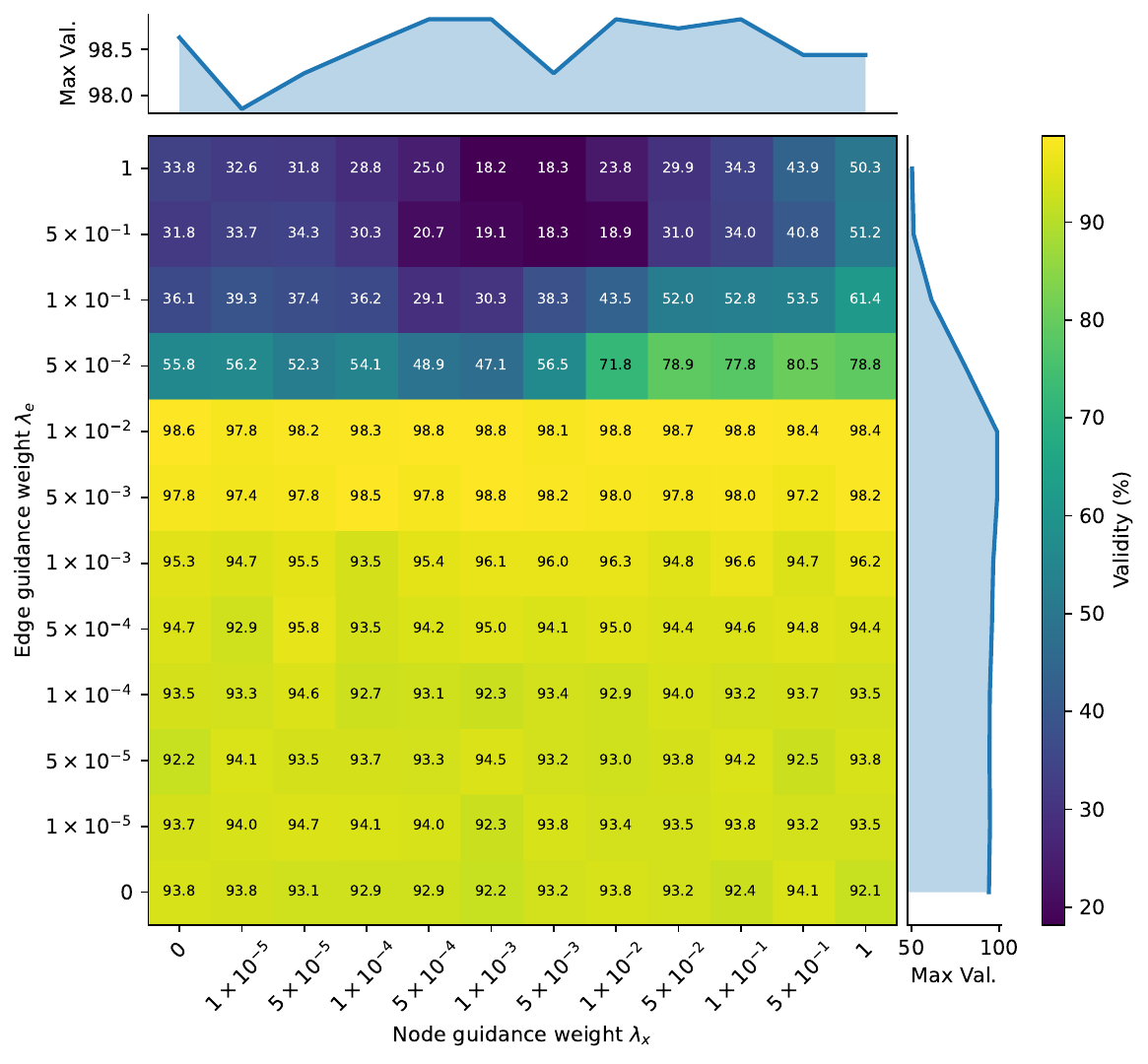}
        \caption{wd-movies}
    \end{subfigure}
    \begin{subfigure}[t]{0.4\textwidth}
        \centering
        \includegraphics[width=\textwidth]{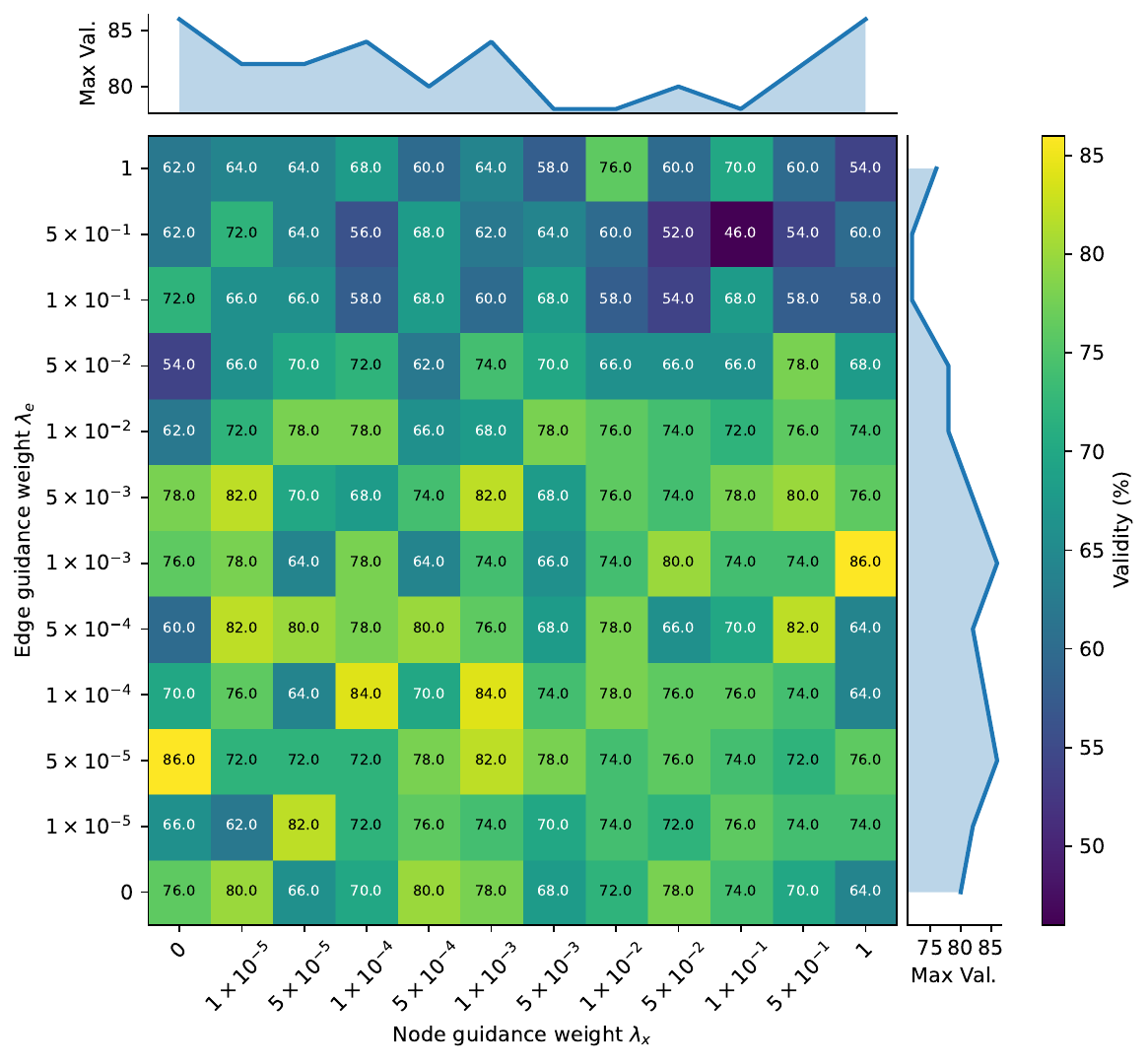}
        \caption{wd-articles}
    \end{subfigure}
    \caption{Grid-search validity (\%) for the LTN-guidance weights $\lambda_x$ and $\lambda_e$ on the five datasets. The color-bar scale is set independently in each panel to emphasize differences within each dataset, so colors should not be compared across datasets.}%
    \label{fig:hyperparameter_heatmaps}
\end{figure*}

In this section we report the per-dataset hyperparameter search we performed for LTN-guidance, the values we chose, implementation and experimental details.
As mentioned in the main text, we split the guidance weight $\lambda$ into node and edge components $\lambda_x$ and $\lambda_e$. We selected their values through a grid search for each dataset. For both hyperparameters, we evaluated the values in the set:
\begin{equation}
    \begin{gathered}
        \{0,10^{-5}, 5\cdot 10^{-5}, 10^{-4}, 5\cdot 10^{-4}, 10^{-3}, \\ 5\cdot 10^{-3}, 10^{-2}, 5\cdot 10^{-2}, 10^{-1}, 5\cdot 10^{-1}, 1\},
    \end{gathered}
\end{equation}
and run the generation of 1000 graphs for all datasets, and 50 graphs for \texttt{wd-articles} for feasibility due to the large size its graphs. All experiments are run with the model obtained from seed 0.
Figure~\ref{fig:hyperparameter_heatmaps} reports the resulting validity for each dataset. We then selected the configuration attaining the maximum validity for each dataset, as summarized in Table~\ref{tab:selected_hyperparameters}.

\begin{table}[t]
    \centering
    \caption{Selected LTN-guidance hyperparameters and corresponding maximum validity (\%).}
    \label{tab:selected_hyperparameters}
    \begin{tabular}{lrrr}
        \toprule
        Dataset & Max. validity & $\lambda_x$ & $\lambda_e$ \\
        \midrule
        syn-paths   & 99.71 & $5\mathbin{\cdot}10^{-4}$ & $10^{-3}$ \\
        syn-tipr    & 99.71 & $5\mathbin{\cdot}10^{-4}$ & $5\mathbin{\cdot}10^{-5}$ \\
        syn-types   & 97.07 & $5\mathbin{\cdot}10^{-3}$ & $1$ \\
        wd-movies   & 98.83 & $5\mathbin{\cdot}10^{-4}$ & $10^{-2}$ \\
        wd-articles & 86.00 & $1$ & $10^{-3}$ \\
        \bottomrule
    \end{tabular}
\end{table}

The implementation uses PyTorch~\cite{ansel2024pytorch}, PyTorch Lightning~\cite{falcon2019pytorch}, Hydra~\cite{yadan2019hydra}, and PyTorch Geometric~\cite{fey2019fast}. All experiments were run on a Linux system using one NVIDIA A40 GPU, five CPU cores, and 20~GB of RAM. The detailed list of architecture hyperparameters, and the list of libraries and their versions are provided in the supplementary code, in the complete Conda environment specification and hydra configurations. The code will be made publicly available upon acceptance of the paper.

\section{Additional Rules for Support Restriction and Shift}

Let $\phi_D$ denote the original conjunction of validity rules for dataset $D$. Then the set of valid graphs is:
\begin{equation}
    \mathcal{M}(\phi_D)=\{G:\phi_D(G)=1\}.
\end{equation}
Then, a \emph{compatible rule} $\psi_D^\text{comp}$ for $D$ is such that:
\begin{equation}
    \mathcal{M}(\phi_D)\cap\mathcal{M}(\psi_D^\text{comp})\neq\emptyset,
\end{equation}
while a \emph{conflicting rule} $\psi_D^\text{conf}$ for $D$ is such that:
\begin{equation}
    \mathcal{M}(\phi_D)\cap\mathcal{M}(\psi_D^\text{conf})=\emptyset.
\end{equation}
Then, a shifted ruleset $\phi_D'$ is obtained from $\phi_D$ such that $\psi_D^\text{conf}$ is not conflicting with it, for example, by removing a rule from $\phi_D$ that is in conflict with $\psi_D^\text{conf}$.
Then, we are interested in sampling, with LTN-guidance, from $p_\theta(G|\phi_D\land\psi_D^\text{comp})$ and $p_\theta(G|\phi_D'\land\psi_D^\text{conf})$, which are the conditional distributions of graphs under \emph{support restriction} and \emph{support shift}, respectively. The former contains a subset of the structures found in the data distribution, while the latter contains graphs out of the data distribution. The objective is, for these models, to maximize the probability of sampling valid graphs. In the main text we reported the results of the experiments for the five datasets, and here we provide the FOL formulas for the additional rules used in the support restriction and shift experiments. For the detailed rule sets of each dataset, see Section 7.4 of the IntelliGraphs paper~\cite{thanapalasingam2023intelligraphs}.

\paragraph{\texttt{syn-paths}.}
For support restriction, we retain the original three-edge path and root rules and require the relations to occur exactly in the order \texttt{cycle\_to}, \texttt{drive\_to}, and \texttt{train\_to} from root to leaf:
\begin{equation}
\begin{aligned}
\exists x,y,z,w\quad &root(x)\land cycle\_to(x,y)\\
&\land drive\_to(y,z)\land train\_to(z,w).
\end{aligned}
\end{equation}
For support shift, we remove the path and root rules and replace them with a directed three-cycle. The three nodes are pairwise distinct:
\begin{equation}
\begin{aligned}
\exists x,y,z\quad &x\neq y\land y\neq z\land z\neq x\\
&\land edge(x,y)\land edge(y,z)\\
&\land edge(z,x).
\end{aligned}
\end{equation}

\paragraph{\texttt{syn-tipr}.}
For support restriction, we retain the canonical five-triple schema and require every tenure to last at least ten years:
\begin{equation}
\begin{aligned}
\forall x,y,z\quad &end\_year(x,y)\land start\_year(x,z)\\
&\rightarrow before\_10\_years(y,z).
\end{aligned}
\end{equation}
Here $y$ and $z$ are year nodes and are identified with their numerical year values in the comparison. For support shift, we retain the five-triple schema but replace the original temporal-order rule with its strict reverse:
\begin{equation}
\begin{aligned}
\forall x,y,z\quad &end\_year(x,y)\land start\_year(x,z)\\
&\rightarrow after(y,z),
\end{aligned}
\end{equation}
meaning that $z$ must come after $y$.

\paragraph{\texttt{syn-types}.}
For support restriction, we retain all original typing rules and the ten-edge constraint, and require the composition of three $spoken\_in$, three $could\_be\_part\_of$, and four $same\_type\_as$ edges:
\begin{equation}
\begin{aligned}
&\phi_{\text{card},3}(spoken\_in)\\
&\land\phi_{\text{card},3}(could\_be\_part\_of)\\
&\land\phi_{\text{card},4}(same\_type\_as)\\
&\land\phi_{\text{card},10}(edge).
\end{aligned}
\end{equation}
For support shift, we replace the ten-edge constraint with twelve edges, using four edges of each relation, while retaining the original typing rules:
\begin{equation}
\begin{aligned}
&\phi_{\text{card},4}(spoken\_in)\\
&\land\phi_{\text{card},4}(could\_be\_part\_of)\\
&\land\phi_{\text{card},4}(same\_type\_as)\\
&\land\phi_{\text{card},12}(edge).
\end{aligned}
\end{equation}

\paragraph{\texttt{wd-movies}.}
For support restriction, we retain all original rules and add the requirement that the movie have at least one genre:
\begin{equation}
\exists x\quad has\_genre(\mathtt{existential\_node},x)
\land genre(x).
\end{equation}
For support shift, we remove the mandatory-director rule, require at least one actor and one genre, and prohibit all $has\_director$ edges:
\begin{equation}
\begin{aligned}
&\exists x\quad has\_actor(\mathtt{existential\_node},x)\land person(x),\\
&\exists x\quad has\_genre(\mathtt{existential\_node},x)\land genre(x),\\
&\forall x,y\;\neg has\_director(x,y).
\end{aligned}
\end{equation}

\paragraph{\texttt{wd-articles}.}
For support restriction, we retain all original rules and require at least one citation from the article to an IRI:
\begin{equation}
\exists x\quad cites(\mathtt{article\_node},x)\land iri(x).
\end{equation}
For support shift, we replace mandatory authorship with mandatory citation and prohibit all $has\_author$ edges:
\begin{equation}
\begin{aligned}
&\exists x\quad cites(\mathtt{article\_node},x)\land iri(x),\\
&\forall x,y\quad\neg has\_author(x,y).
\end{aligned}
\end{equation}
The original subject and citation typing rules and all applicable structural rules are retained.

\section{Knowledge Graphs and Conversion to Labeled Graphs}

Knowledge Graphs (KGs)~\cite{hogan2021knowledge} are graph-based data representations that cover a variety of data from diverse sources. KGs represent information with nodes being \emph{entities} of interest and edges being \emph{relations} between them. This structure allows knowledge to evolve in a flexible manner by enabling granular management of individual entities and relations and allowing missing data.  KGs alone only represent data and an \emph{ontology} may be used to define the schema of the graph, i.e., the types of entities and relations, and the rules they must satisfy. 

An ontology can be implemented by FOL, starting from the definition of a domain of discourse $\domain$ of entities, and from there defining functions and predicates. The domain is partitioned into subsets $\domain_1,\dots,\domain_N$, each identifying an \textit{entity type}. A unary predicate $P_i$ is specified for each subset $\domain_i$. The \textit{relations set} $\relations=\{\relations_1,\dots,\relations_M\}$ contains subsets $\mathcal{R}_j\subseteq\domain\times\domain$. A binary predicate $R_j$ is specified for each relation $\relations_j$. Thus, the triples set becomes
\begin{equation}
    \triples=\{(s,r,o):s,o\in\domain,r\in\relations\}.
\end{equation}

The Resource Description Framework (RDF)~\cite{cyganiak2014rdf} is a standardized model for representing KGs as a set of triples $\triples$, where each triple is of the form $(s,r,o)$, with $s$ being the subject entity, $r$ the edge predicate or relation, and $o$ the object entity, each represented by an identifier.

To define a mapping between KGs in RDF format and labeled graphs, we propose to represent KG entities as node labels, and relations as edge labels. Mathematically, the node labels set is $\labels_v=\domain$ and the edge labels set is $\labels_e=\relations$. To convert the triples set $\triples$ to a labeled graph $G$, we first collect the set of unique entities $\labels(\triples)$ in $\triples$:
\begin{align}
    \labels(\triples)&=\{s \mid (s,r,o)\in\triples\}\cup\{o \mid (s,r,o)\in\triples\}.
\end{align}
Then, we create the vertices $\verts$, and couple them to entities with a bijection $\Lambda_v$:
\begin{align}
    &\verts = \{v_1,\dots,v_n\},&\text{with }n=|\labels(\triples)|\\
    &\Lambda_v(v) = l, & v\in\verts,\ l\in\labels(\triples).
\end{align}
Finally, edges and their labels are defined as:
\begin{align}
    &\edges=\{(v,v') \mid v,v'\in\verts,\ (\Lambda_v(v),r,\Lambda_v(v'))\in\triples\},\\
    &\Lambda_e((v,v'))=r,\quad \text{if}\ (\Lambda_v(v),r,\Lambda_v(v'))\in\triples \land (v,v')\in\edges.
\end{align}
To convert a graph $G$ to triples $\triples$ it is enough to collect all edges and labels into the set:
\begin{equation}
    \triples=\{(\Lambda_v(v),\Lambda_e(v,v'),\Lambda_v(v')) \mid (v,v')\in\edges\}.\label{eq:graph_to_triples}
\end{equation}
Because in principle $\Lambda_v$ is not assumed to be injective, it is possible to have multiple nodes having the same label. In the field of KGs, this implies having two nodes representing the same entity, implying that the set of edges connecting them to other nodes are merged once mapped to triples. Thus, we include additional constraints $\forall l\in\labels_v$:
\begin{equation}
    \forall x,y\ (l(x)\land l(y))\to x=y.
\end{equation}

\ifdefined\ARXIVCOMBINED\else
\bibliography{refs}

\end{document}
\fi

\clearpage
\def\isChecklistMainFile{1}
\makeatletter
\@ifundefined{isChecklistMainFile}{
  \newif\ifreproStandalone
  \reproStandalonetrue
}{
  \newif\ifreproStandalone
  \reproStandalonefalse
}
\makeatother

\ifreproStandalone
\documentclass[letterpaper]{article}
\usepackage[submission]{aaai2027}
\setlength{\pdfpagewidth}{8.5in}
\setlength{\pdfpageheight}{11in}
\frenchspacing

\begin{document}
\fi
\setlength{\leftmargini}{20pt}
\makeatletter\def\@listi{\leftmargin\leftmargini \topsep .5em \parsep .5em \itemsep .5em}
\def\@listii{\leftmargin\leftmarginii \labelwidth\leftmarginii \advance\labelwidth-\labelsep \topsep .4em \parsep .4em \itemsep .4em}
\def\@listiii{\leftmargin\leftmarginiii \labelwidth\leftmarginiii \advance\labelwidth-\labelsep \topsep .4em \parsep .4em \itemsep .4em}\makeatother

\setcounter{secnumdepth}{0}
\renewcommand\thesubsection{\arabic{subsection}}
\renewcommand\labelenumi{\thesubsection.\arabic{enumi}}

\newcounter{checksubsection}
\newcounter{checkitem}[checksubsection]

\newcommand{\checksubsection}[1]{%
  \refstepcounter{checksubsection}%
  \paragraph{\arabic{checksubsection}. #1}%
  \setcounter{checkitem}{0}%
}

\newcommand{\checkitem}{%
  \refstepcounter{checkitem}%
  \item[\arabic{checksubsection}.\arabic{checkitem}.]%
}
\newcommand{\question}[2]{\normalcolor\checkitem #1 #2 \color{blue}}
\newcommand{\ifyespoints}[1]{\makebox[0pt][l]{\hspace{-15pt}\normalcolor #1}}

\section*{Reproducibility Checklist}

\vspace{1em}
\hrule
\vspace{1em}

\textbf{Instructions for Authors:}

This document outlines key aspects for assessing reproducibility. Please provide your input by editing this \texttt{.tex} file directly.

For each question (that applies), replace the ``Type your response here'' text with your answer.

\vspace{1em}
\noindent
\textbf{Example:} If a question appears as
\begin{center}
\noindent
\begin{minipage}{.9\linewidth}
\ttfamily\raggedright
\string\question \{Proofs of all novel claims are included\} \{(yes/partial/no)\} \\
Type your response here
\end{minipage}
\end{center}
you would change it to:
\begin{center}
\noindent
\begin{minipage}{.9\linewidth}
\ttfamily\raggedright
\string\question \{Proofs of all novel claims are included\} \{(yes/partial/no)\} \\
yes
\end{minipage}
\end{center}
Please make sure to:
\begin{itemize}\setlength{\itemsep}{.1em}
\item Replace ONLY the ``Type your response here'' text and nothing else.
\item Use one of the options listed for that question (e.g., \textbf{yes}, \textbf{no}, \textbf{partial}, or \textbf{NA}).
\item \textbf{Not} modify any other part of the \texttt{\string\question} command or any other lines in this document.\\
\end{itemize}

You can \texttt{\string\input} this .tex file right before \texttt{\string\end\{document\}} of your main file or compile it as a stand-alone document. Check the instructions on your conference's website to see if you will be asked to provide this checklist with your paper or separately.

\vspace{1em}
\hrule
\vspace{1em}


\checksubsection{General Paper Structure}
\begin{itemize}

\question{Includes a conceptual outline and/or pseudocode description of AI methods introduced}{(yes/partial/no/NA)}
yes

\question{Clearly delineates statements that are opinions, hypothesis, and speculation from objective facts and results}{(yes/no)}
yes

\question{Provides well-marked pedagogical references for less-familiar readers to gain background necessary to replicate the paper}{(yes/no)}
yes

\end{itemize}
\checksubsection{Theoretical Contributions}
\begin{itemize}

\question{Does this paper make theoretical contributions?}{(yes/no)}
no

	\ifyespoints{\vspace{1.2em}If yes, please address the following points:}
        \begin{itemize}
	
	\question{All assumptions and restrictions are stated clearly and formally}{(yes/partial/no)}
	NA

	\question{All novel claims are stated formally (e.g., in theorem statements)}{(yes/partial/no)}
	NA

	\question{Proofs of all novel claims are included}{(yes/partial/no)}
	NA

	\question{Proof sketches or intuitions are given for complex and/or novel results}{(yes/partial/no)}
	NA

	\question{Appropriate citations to theoretical tools used are given}{(yes/partial/no)}
	NA

	\question{All theoretical claims are demonstrated empirically to hold}{(yes/partial/no/NA)}
	NA

	\question{All experimental code used to eliminate or disprove claims is included}{(yes/no/NA)}
	NA
	
	\end{itemize}
\end{itemize}

\checksubsection{Dataset Usage}
\begin{itemize}

\question{Does this paper rely on one or more datasets?}{(yes/no)}
yes

\ifyespoints{If yes, please address the following points:}
\begin{itemize}

	\question{A motivation is given for why the experiments are conducted on the selected datasets}{(yes/partial/no/NA)}
	yes

	\question{All novel datasets introduced in this paper are included in a data appendix}{(yes/partial/no/NA)}
	NA

	\question{All novel datasets introduced in this paper will be made publicly available upon publication of the paper with a license that allows free usage for research purposes}{(yes/partial/no/NA)}
	NA

	\question{All datasets drawn from the existing literature (potentially including authors' own previously published work) are accompanied by appropriate citations}{(yes/no/NA)}
	yes

	\question{All datasets drawn from the existing literature (potentially including authors' own previously published work) are publicly available}{(yes/partial/no/NA)}
	yes

	\question{All datasets that are not publicly available are described in detail, with explanation why publicly available alternatives are not scientifically satisficing}{(yes/partial/no/NA)}
	NA

\end{itemize}
\end{itemize}

\checksubsection{Computational Experiments}
\begin{itemize}

\question{Does this paper include computational experiments?}{(yes/no)}
yes

\ifyespoints{If yes, please address the following points:}
\begin{itemize}

	\question{This paper states the number and range of values tried per (hyper-) parameter during development of the paper, along with the criterion used for selecting the final parameter setting}{(yes/partial/no/NA)}
	yes

	\question{Any code required for pre-processing data is included in the appendix}{(yes/partial/no)}
	yes

	\question{All source code required for conducting and analyzing the experiments is included in a code appendix}{(yes/partial/no)}
	yes

	\question{All source code required for conducting and analyzing the experiments will be made publicly available upon publication of the paper with a license that allows free usage for research purposes}{(yes/partial/no)}
	yes
        
	\question{All source code implementing new methods have comments detailing the implementation, with references to the paper where each step comes from}{(yes/partial/no)}
	yes

	\question{If an algorithm depends on randomness, then the method used for setting seeds is described in a way sufficient to allow replication of results}{(yes/partial/no/NA)}
	yes

	\question{This paper specifies the computing infrastructure used for running experiments (hardware and software), including GPU/CPU models; amount of memory; operating system; names and versions of relevant software libraries and frameworks}{(yes/partial/no)}
	yes

	\question{This paper formally describes evaluation metrics used and explains the motivation for choosing these metrics}{(yes/partial/no)}
	yes

	\question{This paper states the number of algorithm runs used to compute each reported result}{(yes/no)}
	yes

	\question{Analysis of experiments goes beyond single-dimensional summaries of performance (e.g., average; median) to include measures of variation, confidence, or other distributional information}{(yes/no)}
	yes

	\question{The significance of any improvement or decrease in performance is judged using appropriate statistical tests (e.g., Wilcoxon signed-rank)}{(yes/partial/no)}
	no

	\question{This paper lists all final (hyper-)parameters used for each model/algorithm in the paper’s experiments}{(yes/partial/no/NA)}
	yes

\end{itemize}
\end{itemize}
\ifreproStandalone
\end{document}
\fi

\end{document}